\documentclass[sn-mathphys,Numbered]{sn-jnl}

\usepackage{graphicx}%
\usepackage{subfig}%
\usepackage{multirow}%
\usepackage{amsmath,amssymb,amsfonts}%
\usepackage{amsthm}%
\usepackage{mathrsfs}%
\usepackage[title]{appendix}%
\usepackage{xcolor}%
\usepackage{textcomp}%
\usepackage{tabularx}
\usepackage{manyfoot}%
\usepackage{booktabs}%
\usepackage{algorithm}%
\usepackage{algorithmicx}%
\usepackage{hyperref}%
\usepackage{array,ragged2e,longtable}%
\usepackage{algpseudocode}%
\usepackage{listings}%
\usepackage{adjustbox}%
\usepackage{geometry}
\usepackage{adjustbox}
\newcolumntype{P}{>{\RaggedRight\hspace{0pt}}p{\dimexpr(\textwidth-1cm -10\tabcolsep -5\arrayrulewidth)/4\relax}}

\theoremstyle{thmstyleone}%
\theoremstyle{thmstyletwo}%

\theoremstyle{thmstylethree}%

\begin{document}

\title[Article Title]{Graph Neural Network approaches for single-cell data: A recent overview.}


\author*[1]{\fnm{Konstantinos} \sur{Lazaros}} \email{std528087@ac.eap.gr}

\author[1]{\fnm{Dimitris E.} \sur{Koumadorakis}}

\author[1]{\fnm{Panagiotis} \sur{Vlamos}}

\author[1]{\fnm{Aristidis G.} \sur{Vrahatis}}

\affil*[1]{\orgdiv{Bioinformatics and Human Electrophysiology Laboratory, Department of Informatics}, \orgname{Ionian University}, \orgaddress{\city{Corfu}, \postcode{49100}, \country{Greece}}}








\abstract{Graph Neural Networks (GNN) are reshaping our understanding of biomedicine and diseases by revealing the deep connections among genes and cells. As both algorithmic and biomedical technologies have advanced significantly, we're entering a transformative phase of personalized medicine. While pioneering tools like Graph Attention Networks (GAT) and Graph Convolutional Neural Networks (Graph CNN) are advancing graph-based learning, the rise of single-cell sequencing techniques is reshaping our insights on cellular diversity and function. Numerous studies have combined GNNs with single-cell data, showing promising results. In this work, we highlight the GNN methodologies tailored for single-cell data over the recent years. We outline the diverse range of graph deep learning architectures that center on GAT methodologies. Furthermore, we underscore the several objectives of GNN strategies in single-cell data contexts, ranging from cell-type annotation, data integration and imputation, gene regulatory network reconstruction, clustering and many others. This review anticipates a future where GNNs become central to single-cell analysis efforts, particularly as vast omics datasets are continuously generated and the interconnectedness of cells and genes enhances our depth of knowledge in biomedicine.}

\keywords{scRNA-seq, spatial transcriptomics, graph neural networks, graph attention}

\maketitle

\section{Introduction}

Single-cell sequencing, is a cutting-edge next-generation sequencing technique, that enables intricate analysis of individual cell genomes or transcriptomes, illuminating heterogeneity in cellular populations. Unlike traditional sequencing methods which present an averaged view of numerous cells, thereby obscuring finer distinctions and nuances, single-cell sequencing zeroes in on the unique genomic and transcriptomic expressions of each cell. This unparalleled precision exposes the extent and variability of gene expression within specific cell populations, spotlighting the vast behavioral, functional, and structural diversities—collectively termed heterogeneity—originating from varying gene expression patterns \cite{tang2019single}. 

Complementing this, recent innovations in next-generation sequencing and imaging have birthed the technique of spatial transcriptomics. This powerful method systematically maps gene expression across tissue spaces, a pivotal advance that has garnered significant insights in fields such as neuroscience, developmental biology, plant biology, and oncology. Notably, the clinicians recognize the immense diagnostic value of spatial organization within tissue—often discerned through histopathology—as many diseases manifest as aberrations in this spatial matrix \cite{rao2021exploring}.

Network medicine serves as an advanced, unbiased platform, assimilating multiomics data from diverse biological levels to elevate the precision and therapeutic approach for types of disease (such as cardivascular diseases) \cite{wang2023multiomics}. The inherent strength of such multi-omics network medicine strategies is manifested in its capability to pinpoint, and dissect the heterogeneity inherent in many compounded biological phenomena. This nuanced understanding not only highlights the intricacies of complex conditions but also steers the direction of customized drug therapies, setting the stage for a new era of precision medicine. In this context, biological systems can be conceptualized as networks of nodes and edges. The nodes represent diverse entities, ranging from genes, proteins, and metabolites among others, while the edges represent their interactions \cite{blencowe2019network}. Given the rapid expansion and heterogeneity of multi-omics data, there's been a corresponding surge in the development of accurate biological networks as well as reliable tools and methodologies that will be used for effective and information-rich analyses.

Graph theory, is a profound mathematical discipline focusing on the study of graphs, that has ascended to prominence as an indispensable tool for analyzing intricate systems and their relations \cite{gnn_foundations}. Within this context, a graph is depicted as an ensemble of nodes (or vertices) interconnected by edges, representing the relationships between entities. Such a representational system facilitates the interpretation of complex networks, enabling profound insights into their inherent structures and recurring patterns. The efficacy of machine learning approaches is contingent not only upon the meticulous design of the encompassing algorithms but also on the quality and fidelity of data representation. Suboptimal representations, either devoid of pivotal details or plagued by superfluous or erroneous information, could undermine the algorithm's operational efficiency across various tasks. Representation learning endeavors to filter data in order to capture sufficient and at the same time minimal information. Recently, network representation learning (NRL) has piqued considerable research interest. The goal of NRL is to derive latent, dimensionally reduced representations of network vertices, whilst preserving the network's topological integrity, vertex attributes, and other information. Once these transformed vertex representations are procured, ensuing network analytical tasks can be easily executed, relying on traditional vector-based machine learning frameworks within this transformed latent space. 

Deep learning has become a staple within the fields of artificial intelligence and machine learning, demonstrating unparalleled efficacy in domains such as image processing, and natural language processing. While graphs can undeniably be applied to myriad real-world scenarios, harnessing deep learning for graph data remains a complex endeavor. This complexity stems from various factors such as the inherent non-uniformity of graph structures, contrasting with the regularity of grid structured data, such as images or audio. For instance establishing operations like convolution and pooling, quintessential to convolutional neural networks (CNNs), for graphs is not a simple task \cite{georgousis2021graph}. The multifaceted nature of graphs, encompassing diverse attributes and characteristics, which necessitates different model architectures. The sheer magnitude of modern graphs, often encompassing millions or even billions of nodes and edges, prompting a dire need for scalable models. The interdisciplinary integration of graphs with domains like biology, chemistry, and social sciences. While this combination offers opportunities by leveraging domain-specific knowledge, it simultaneously increases the complexity of model conception. 

In contemporary research, graph neural networks (GNNs) have garnered significant attention. The architectures and strategies employed exhibit vast heterogeneity, spanning from supervised to unsupervised ones, and from convolutional frameworks to recursive ones, inclusive of architectures like graph recurrent neural networks (Graph RNNs) \cite{ruiz2020gated}, graph convolutional networks (GCNs) \cite{kipf2016semi}, graph autoencoders (GAEs) \cite{kipf2016variational}, graph reinforcement learning (Graph RL) \cite{mingshuo2022reinforcement}, graph adversarial methodologies \cite{chen2020survey} as well as graph attention networks (GATs) \cite{velivckovic2017graph}. To elaborate further, Graph RNNs discern recursive and sequential graph patterns by modeling states at the level of nodes or the graph. GCNs establish convolutional and readout processes on non-uniform graph matrices, capturing common local and global patterns. On the other hand GAEs presuppose low-rank graph structures and employ unsupervised techniques for node representation. Moreover, Graph RL contextualizes graph-based actions and feedback, adhering to preset constraints. Graph adversarial methodologies employ adversarial training strategies to augment the generalization potential of graph-based models, evaluating their resilience against adversarial attacks. Last but not least graph attention networks (GATs) are in essence an improvement over graph convolutional networks. They are built upon the concept of self-attention a process that has been popularized by transformer based models such as BERT and GPT-3 \cite{hands_on_gnn}. Through this mechanism the importance of each node in the graph is learned dynamically instead of explicitly assigned as in GCNs. Such attention mechanisms are being utilized for a wide variety of tasks such as natural language undestanding and computer vision. 

Recent breakthroughs in multimodal single-cell technologies have ushered in an era where simultaneous acquisition of diverse omics data from individual cells is feasible, offering profound insights into cellular states and dynamics. Nonetheless, crafting joint representations from such multimodal datasets, identifying the intricacies between modalities, and most crucially, assimilating the plethora of unimodal datasets for subsequent analysis, remain formidable challenges. Graph neural network-based frameworks present a promising avenue to tackle these obstacles, thereby streamlining the analysis of multimodal single-cell data. Over the past three years, numerous GNN-based tools have been proposed for various types of single-cell analyses, the latest of which will be delved into in ensuing sections. \textbf{Fig. 1} offers an overview of the functioning of GNN-based tools in the context of single-cell data analysis.

\vspace{1mm}

\begin{figure}[!htbp]
    \centering
    \begin{adjustbox}{width=\textwidth,center}
        \includegraphics[width=\textwidth]{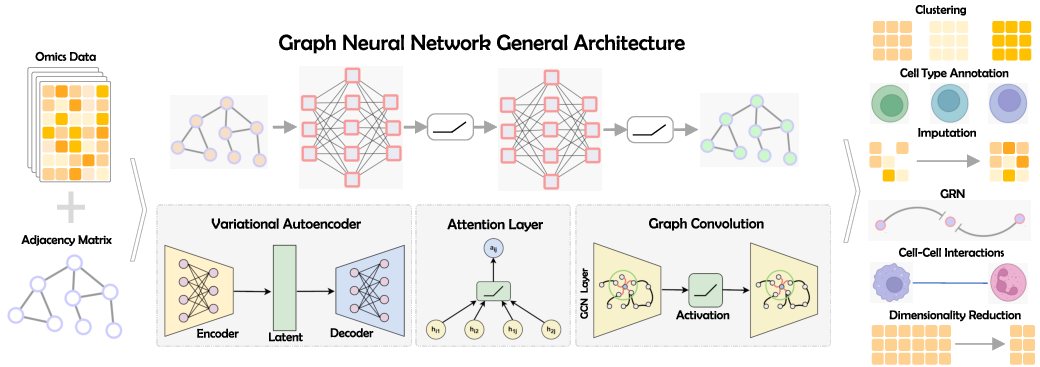}
    \end{adjustbox}
    \caption{The figure delineates a representative pipeline for single-cell sequencing data analysis via Graph Neural Networks (GNNs). On the left, one observes the typical inputs for the GNN architecture, primarily encompassing a single-cell omics expression matrix and a graph, depicted here as an adjacency matrix. This graph could encapsulate interactions of various kinds, including but not limited to cell-cell or gene-gene interactions. The central upper segment illustrates a generalized GNN framework, purposed for the generation of refined vector representations of the input data. Several GNN architectures, highlighted in the central lower portion, such as graph variational autoencoders, graph attention networks, and graph convolutional networks, can be employed to this end. As articulated on the right, these vector representations, once crafted by the GNN, can be further leveraged for numerous subsequent analytical tasks in single-cell analysis. These tasks include, but are not limited to, cell-cell interaction inference, cell-type annotation, imputation of absent values, Gene Regulatory Network (GRN) inference, cell clustering, and dimensionality reduction.}
    \label{fig:your_label}
\end{figure}

\section{Recent GNN-based single-cell-sequencing analysis tools}
Over the past four years an impressive amount of GNN-based tools has emerged proposed for various tasks that are staples of single-cell sequencing data downstream analyses, such as imputation, clustering and cell annotation/identification among others. In this section and the following subsections we are going to explore the forefront of GNN-based single-cell data analysis, delving into 39 cutting edge computational methods that aid in highlighting and discerning the intricacies inherent to cellular heterogeneity. These tools have been implemented for various parts of single-cell omics data analysis including but not limited to imputation, clustering, dimensionality reduction, cell annotation, GRN inference and cell-cell interaction inference among others. Through analyzing these state-of-the-art we are able to obtain an extensive understanding of recent progress taking place in graph-based methods for single-cell analysis. Details about the tools discussed in this work are summarized in \textbf{Table 1}.

\subsection{Imputation}

A major characteristic of scRNA-seq data is their pronounced sparsity, manifested as a heightened fraction of observed "zeros." These zeros, representing instances where no unique molecular identifiers (UMIs) or reads map to a specific gene in a cell, pose significant challenges for accurate downstream analyses. Addressing this sparsity is of paramount importance in order to harness the true potential of scRNA-seq data. In this context, "imputation" has surfaced as a salient approach. Imputation techniques, drawing parallels from the genomics realm where missing or unobserved genotype data is estimated, endeavor to infer the most plausible values in place of these zeros, thereby providing a more comprehensive representation of the cellular transcriptome \cite{hou2020systematic}. However, conventional machine learning imputation strategies have been observed to grapple with the intricacies of scRNA-seq data, particularly its non-linear associations and unique counting architectures. This inadequacy underscores the necessity for more advanced methodologies \cite{rao2021imputing}. Among the myriad of solutions, graph deep learning-based imputation has emerged as particularly promising. Such models, leveraging graph neural networks, are adept at discerning the intricate relationships between cells, enabling them to impute missing data with heightened precision.

In this vein, ScGAEGAT \cite{feng2022single} is a multi-modal model integrating graph autoencoders and graph attention networks, specifically tailored for scRNA-seq analysis through graph neural networks. Its gene imputation performance is measured using metrics such as cosine similarity, median L1 distance, and root-mean-squared error, whereas its cell clustering capability is evaluated using adjusted mutual information, normalized mutual information, completeness score, and Silhouette coefficient score. Designed on a foundation of multi-modal graph autoencoders (GAEs) and graph attention (GAT) networks, scGAEGAT adeptly models heterogeneous cell-cell interactions and intricate gene expression patterns inherent in scRNA-seq data. Operating through the use of an encoder-decoder deep learning framework, the model not only facilitates gene imputation and cell clustering but also offers a comprehensive vantage point for probing cellular interactions by discerning associations across the entire cell population.

The architecture of scGAEGAT involves the incorporation of a pre-treatment gene expression matrix, as input to a feature autoencoder. This component constructs and subsequently refines the cell graph using acquired embeddings. By infusing a graph attention mechanism into the graph autoencoder, the model takes the cell graph that has been created as input and assigns variable weights to individual nodes, enhancing its capacity to delineate cellular relationships and achieve precise cell-type clustering. In this sophisticated setup, each cell type benefits from a dedicated cluster autoencoder to reconstruct gene expression values. These reconstructed values are subsequently looped back as fresh inputs in iterative rounds until a point of convergence is achieved. Central to scGAEGAT's efficacy is the GAT mechanism, granting the model the ability to allocate differential weights to cells via the GAT coefficient. Throughout its iterative operations, the model is composed of four robust autoencoders: a feature autoencoder addressing gene expression regulation, a graph autoencoder establishing cell-cell interactions, a cluster autoencoder pinpointing cell type, and an imputation autoencoder dedicated to the recovery of the gene expression matrix. By harnessing topological insights between genes, scGAEGAT conveys cell-cell interactions via a low-dimensional embedding derived from multiple autoencoder structure training vectors. This mechanism proves pivotal for cell type aggregation, trajectory-based cell arrangement inference, and the imputation of any absent data to enhance gene correlations.

Similarly, envisioned as a dropout imputation method, GNNImpute \cite{xu2021efficient} is an advanced autoencoder network that leverages graph attention convolution to aggregate multi-level cell similarity information, utilizing convolution operations on non-Euclidean space for scRNA-seq data. Unique from existing imputation methods, this model proficiently imputes dropouts while simultaneously mitigating dropout noise. Its performance is evaluated using metrics like mean square error (MSE), mean absolute error (MAE), Pearson correlation coefficient (PCC), and Cosine similarity (CS). Through constructing the scRNA-seq data graph, GNNImpute employs a graph attention convolutional layer to discern and select similar adjacent nodes. Following this, it aggregates these analogous neighboring nodes. Nodes within this graph can perpetually relay messages in the direction along the edges until a stability point emerges. Through this mechanism, GNNImpute facilitates the embedding of cellular expressions from identical tissue regions into low-dimensional vectors via its autoencoder architecture. Not only does it discern co-expression trends among similar cells, but it also filters sequencing technical noise from dropout imputation, enhancing subsequent scRNA-seq data analysis.

A salient feature of GNNImpute is its incorporation of an attention mechanism, which enables weight assignment to different cells based on attention coefficients. This approach fosters the establishment of nonlinear relationships between genes by learning low-dimensional embeddings of expressions through its autoencoder network. In contrast to tools like DCA, GNNImpute possesses the capability to discern gene co-expression patterns among similar cells by aggregating information from multi-level neighbors. The structure of GNNImpute comprises both an encoder and a decoder. The encoder encompasses two graph attention convolutional layers, focusing on the transmission of neighboring node information. Conversely, the decoder is composed of two linear layers. GNNImpute utilizes a masked expression matrix as its model input, and its output serves as the basis for loss value computation, with model parameters subsequently being optimized by this derived value.

Moreover, ScDGAE \cite{feng2023single} is a sophisticated directed graph neural network designed for scRNA-seq analysis, integrating both graph autoencoders and a graph attention network. Distinguished from traditional models, directed graph neural networks not only uphold the intrinsic connection attributes of the directed graph but also enhance the receptive field of convolution operations. To gauge its proficiency in gene imputation, scDGAE is evaluated using metrics such as cosine similarity, median L1 distance, and root-mean-squared error. Meanwhile, its cell clustering capabilities are assessed with tools like adjusted mutual information, normalized mutual information, completeness score, and Silhouette coefficient score.

scDGAE adeptly models heterogeneous cell-cell associations and the gene expression patterns inherent in scRNA-seq data. Beyond merely facilitating gene imputation and cell clustering, scDGAE presents an encoder-decoder deep learning architecture optimized for scRNA-Seq analysis. This innovative framework offers a comprehensive view, capturing the intricate relationships spanning the entire cell population.

The scGNN 2.0 \cite{gu2022scgnn} model is composed of a triad of stacked autoencoders, preceded by a program initialization. Within this phase, the data undergoes preprocessing, and a left-truncated mixture Gaussian (LTMG) model is applied. This statistical model adeptly discerns regulatory signals for individual genes, which subsequently serve as regularization criteria within the feature autoencoders. Contrary to the approach in scGNN 1.0, where imputation is contingent on a distinct imputation autoencoder activated post-iteratively, this supplementary autoencoder has been avoided in the 2.0 version, retaining the three autoencoders. Within the graph autoencoder phase, a multi-head graph attention mechanism is harnessed to calculate a graph embedding.

Upon conclusion of the final iteration, scGNN 2.0 creates: (i) an imputed gene expression matrix in csv format, structured with cells as rows and genes as columns; (ii) a graph embedding matrix in csv format, arranged with cells as rows and the graph embedding dimensions as columns; (iii) a cell graph in csv format as an edge list, with three columns constituting the starting node, the terminal node, and the edge weight; as well as (iv) a list in csv format, containing cell cluster labels, with the first column detailing cell names and the subsequent column indicating cell labels.

Another GNN-based tool for single-cell data imputation is GE-Impute \cite{wu2022ge}, which has designed to address the dropout zeros in scRNA-seq data through a graph embedding-based neural network model. By acquiring a neural graph representation for every cell, GE-Impute effectively reconstructs the cell–cell similarity network. This reconstruction subsequently offers enhanced imputation of dropout zeros, drawing from more precisely designated neighbors within the similarity network. When analyzing the correlation in gene expression between baseline expression data and its simulated dropout counterpart, GE-Impute demonstrates a notably superior efficacy in retrieving dropout zeros for both droplet- and plate-based scRNA-seq data sets. 
GE-Impute commences with the construction of a cell-cell similarity network predicated on Euclidean distance metrics. For each cell, a random walk of a specified length is simulated utilizing both BFS (Breadth-First Search) and DFS (Depth-First Search) strategies. Subsequent to this, a graph embedding-based neural network model is tasked with refining the embedding matrix corresponding to each individual cell, predicated on the sample walks. Leveraging the newly-trained embedding matrix, the similarity metrics among cells are recalculated, facilitating the prediction of novel link-neighbors and the reconstruction of the cell-cell similarity network. Conclusively, GE-Impute carries out the imputation of dropout zeros for each cell, achieved by computing the mean expression value of its associated neighbors in the reproduced similarity network.

\newpage
\newgeometry{top=1cm, bottom=4cm}
\renewcommand*\thefootnote{\alph{footnote}}
\begingroup 
\setlength\extrarowheight{1.5pt}
\small
\setlength\LTcapwidth\textwidth
\setlength\LTleft{-2.5cm}
\begin{longtable}{p{2.5cm} p{2cm} p{4.5cm} p{2.5cm} p{4cm} p{0.5cm} p{1cm} c }

\caption{Recent GNN approaches for single-cell data} \label{BT2} \\
\hline

 \textbf{Name} 
  & \textbf{Graph}
  & \textbf{Input}                                                 
  & \textbf{Architecture}
  & \textbf{Aim}
  & \textbf{Ref}
  \\ \hline \hline
\endfirsthead 

\hline

\multicolumn{5}{r@{}}{\em Cont'd on following page}\\
\endfoot

\endlastfoot


\href{https://github.com/zpliulab/GENELink}{GENELINK} 
  & Tf-gene 
  & \raggedright scRNA-seq + Adjacency Matrix 
  & GAT \footnote{GAT: Graph Attention}
  & GRN 
  & \cite{chen2022graph}
\\ \hline

\href{https://github.com/davidbuterez/CellVGAE}{CELLVGAE} 
  & Cell-cell 
  & scRNA-seq
  & \raggedright VGAE \footnote{VGAE: Variational Graph Auto Encoder} + GAT  
  & Dimensionality Reduction 
  & \cite{buterez2022cellvgae}
\\ \hline

ScGAEGAT 
  & Cell-cell 
  & scRNA-seq
  & \raggedright GAE + GAT
  & Imputation
  & \cite{feng2022single}
\\ \hline

\href{https://github.com/Lav-i/GNNImpute}{GNNImpute}
& Cell-cell
& scRNA-seq
& GATC \footnote{GATC: Graph Attention Convolution}
& Imputation
& \cite{xu2021efficient}
\vspace{-1cm}
\\ \hline

\href{https://github.com/compbiolabucf/omicsGAT}{omicsGAT}
& Cell-cell
& scRNA-seq
& GAT
& \raggedright Disease prediction 
& \cite{baul2022omicsgat}
\vspace{-1cm}
\\ \hline

\href{https://github.com/Joye9285/scGAC}{scGAC}
& Cell-cell
& scRNA-seq
& GATE \footnote{GATE: Graph Attention Auto Encoder}
& Clustering
& \cite{cheng2022scgac}
\vspace{-1cm}
\\ \hline

\href{https://github.com/huoyuying/SpaDAC}{SpaDAC}
& Cell-cell
& Spatial
& GAT
& Clustering
& \cite{huo2023integrating}
\vspace{-1cm}
\\ \hline

\href{https://github.com/zhanglabtools/STAGATE}{STAGATE}
& Cell-cell
& Spatial
& GATE
& \raggedright Spatial domain detection
& \cite{dong2022deciphering}
\vspace{-1cm}
\\ \hline

ScDGAE
& Cell-cell
& scRNA-seq
& GAT
& Imputation
& \cite{feng2023single}
\vspace{-1cm}
\\ \hline
\href{https://github.com/SAkbari93/SCEA}{SCEA}
& Cell-cell
& scRNA-seq
& \raggedright GATE
& \raggedright Cell annotation
& \cite{abadi2023optimized}
\vspace{-1cm}
\\ \hline
\href{https://github.com/OSU-BMBL/scGNN2.0}{scGNN}
& Cell-cell
& scRNA-seq
& GNN \footnote{GNN: Graph Neural Network}
& Imputation
& \cite{gu2022scgnn}
\vspace{-1cm}
\\ \hline

\href{https://github.com/ZixiangLuo1161/scGAE}{scGAE}
& Cell-cell
& scRNA-seq
& GAE \footnote{GAE: Graph Auto Encoder}
& Dimensionality Reduction
& \cite{luo2021topology}
\vspace{-1cm}
\\ \hline

\href{https://github.com/ZzzOctopus/scASGC}{scASGC}
& Cell-cell
& scRNA-seq
& \raggedright GCN + GAT
& Clustering
& \cite{wang2023scasgc}
\vspace{-1cm}
\\ \hline

\href{https://github.com/WHY-17/SCDRHA}{SCDRHA}
& Cell-cell
& scRNA-seq
& \raggedright DCA + GAT
& Dimensionality Reduction
& \cite{zhao2021scdrha}
\vspace{-1cm}
\\ \hline

\href{https://github.com/bhklab/GraphComm}{GraphComm}
& Cell-cell
& \raggedright scRNA-seq + LR data \footnote{LR: Ligand-Receptor}
& GAT
& Cell-cell Interactions
& \cite{so2023graphcomm}
\vspace{-1cm}
\\ \hline

\href{https://github.com/DisscLab/HNNVAT}{HNNVAT}
& Cell-cell
& scRNA-seq
& GCN
& \raggedright Cell annotation
& \cite{wang2023adversarial}
\vspace{-1cm}
\\ \hline

\href{https://github.com/jashahir/cellograph}{cellograph}
& Cell-cell
& scRNA-seq
& GCN
& \raggedright DEGs + visualization
& \cite{shahir2023cellograph}
\vspace{-1cm}
\\ \hline

\href{https://github.com/OSU-BMBL/deepmaps}{DeepMAPs}
& Cell-gene
& scMulti-omics
& HGT \footnote{HGT: Heterogenous Graph Transformer}
& GRN
& \cite{ma2023single}
\vspace{-1cm}
\\ \hline

\href{https://github.com/YangLabHKUST/STitch3D}{Stitch3D}
& Spot-Spot
& \raggedright scRNA-seq + spatial 
& \raggedright Encoder + GAT
& 3d structure from 2d slices
& \cite{wang2023construction}
\vspace{-1cm}
\\ \hline

\href{https://github.com/QSong-github/spaCI}{SpaCI}
& Cell-cell
& Spatial
& \raggedright GAT
& Cell-cell Interactions
& \cite{tang2023spaci}
\vspace{-1cm}
\\ \hline

\href{https://github.com/wxbCaterpillar/GE-Impute}{GE-Impute}
& Cell-cell
& scRNA-seq
& GNN
& Imputation
& \cite{wu2022ge}
\vspace{-1cm}
\\ \hline

\href{https://github.com/JieZheng-ShanghaiTech/PIKE-R2P}{PIKE-R2P}
& PPI
& scRNA-seq
& GNN
& PPI Inference
& \cite{dai2021pike}
\vspace{-1cm}
\\ \hline

GLAE
& Cell-cell
& scRNA-seq
& GNN
& Clustering
& \cite{shan2023glae}
\vspace{-1cm}
\\ \hline

\href{https://github.com/Philyzh8/scTAG}{scTAG}
& Cell-cell
& scRNA-seq
& GCN
& Clustering
& \cite{yu2022zinb}
\vspace{-1cm}
\\ \hline

\href{https://github.com/ZJUFanLab/scDeepSort}{scDeepSORT}
& Cell-gene
& scRNA-seq
& GNN
& \raggedright Cell annotation
& \cite{shao2020reference}
\vspace{-1cm}
\\ \hline

\href{https://github.com/Junseok0207/scGPCL}{scGPCL}
& Cell-gene
& scRNA-seq
& GNN
& Clustering
& \cite{lee2023deep}
\vspace{-1cm}
\\ \hline

\href{https://github.com/changwn/scFEA}{scFEA}
& Metabolomics
& scRNA-seq
& GNN
& \raggedright Flux Estimation
& \cite{alghamdi2021graph}
\vspace{-1cm}
\\ \hline

\href{https://github.com/ericlin1230/scGMM-VGAE}{ScGMM-VGAE}
& Cell-cell
& scRNA-seq
& GMM \footnote{GMM: Gaussian Mixture Model} + VGAE
& Clustering
& \cite{lin2023scgmm}
\vspace{-1cm}
\\ \hline

\href{https://github.com/anlingUA/scAGN}{scAGN}
& Cell-cell
& scRNA-seq
& AGN \footnote{AGN: Attention-based Graph Network}
& \raggedright Cell annotation
& \cite{bhadani2023attention}
\vspace{-1cm}
\\ \hline

\href{https://github.com/ddb-qiwang/scMRA-torch}{scMRA}
& \raggedright Cell type
& scRNA-seq
& GCN
& \raggedright Cell annotation
& \cite{yuan2022scmra}
\vspace{-1cm}
\\ \hline

\href{https://github.com/KrishnaswamyLab/scgraph}{scGraph}
& Gene-gene
& \raggedright scRNA-seq + GIN
& GNN
& \raggedright Cell annotation
& \cite{yin2022scgraph}
\vspace{-1cm}
\\ \hline

\href{https://github.com/gao-lab/GLUE}{GLUE}
& Omics
& scMulti-omics
& VGAE  
& Integration
& \cite{cao2022multi}
\vspace{-1cm}
\\ \hline

\href{https://github.com/sunyolo/DeepTFni}{DeepTFni}
& Tf-gene
& scATAC-seq
& GNN
& GRN
& \cite{li2022inferring}
\vspace{-1cm}
\\ \hline

\href{https://github.com/QSong-github/scGCN}{scGCN}
& Cell-cell
& scRNA-seq
& GCN
& Label Transfer
& \cite{song2021scgcn}
\vspace{-1cm}
\\ \hline

scHCA
& Cell-cell
& \raggedright scRNA-seq + Epigenomics
& GNN
& Cell-cell interactions
& \cite{liu2022heterogeneous}
\vspace{-1cm}
\\ \hline

\href{https://github.com/alexw16/gridnet}{GrID-Net}
& Cell-cell
& scMulti-omics
& GNN
& \raggedright Locus-gene associactions
& \cite{wu2023econometric}
\vspace{-1cm}
\\ \hline

\href{https://github.com/xiaoyeye/GCNG}{GCNG}
& Cell-cell
& Spatial
& GCN
& LR pair identification
& \cite{yuan2020gcng}
\vspace{-1cm}
\\ \hline

\href{https://github.com/luoyuanlab/stdgcn}{StdGCN}
& Spot-Spot
& \raggedright scRNA-seq + spatial 
& GCN
& \raggedright Cell Type Deconvolution
& \cite{li2023spatial}
\vspace{-1cm}
\\ \hline

\href{https://github.com/zcang/SCAN-IT}{SCAN-IT}
& \raggedright Cell-cell
& Spatial
& GNN
& Segmentation
&\cite{cang2021scan}
\vspace{-1cm}
\\ \hline

\end{longtable}
\clearpage

\restoregeometry

\subsection{Clustering}

Single-cell RNA-sequencing (scRNA-seq) has become instrumental in unraveling cellular heterogeneity, necessitating robust analytical approaches. Central to this analytical arsenal is unsupervised clustering, a pivotal method deployed to discern and classify putative cell types from the plethora of information embedded in scRNA-seq datasets. Despite the notable advancements in clustering algorithms witnessed in recent times, the domain remains riddled with ambiguities. Specifically, the scientific community grapples with the conundrum of delineating the most efficacious approach and establishing definitive criteria for cell type definition based on scRNA-seq readings \cite{kiselev2019challenges}. Compounding these challenges is the staggering evolution of sequencing technologies, leading to an exponential growth in the volume of single-cell sequencing data. Classical clustering methodologies are increasingly proving inept at managing large-scale data, struggling to maintain efficiency and grappling with the complexities inherent to single-cell datasets \cite{feng2023single}. In light of these challenges, graph neural network-based tools emerge as a beacon of promise. By exploiting the inherent network structures within cellular data, these models are adept at navigating the vast and intricate scRNA-seq landscapes.

Based on the above ScGAC, (Single-cell Graph Attentional Clustering) \cite{cheng2022scgac}, emerges as an unsupervised clustering model devised specifically for scRNA-seq data analysis. The methodology's initial steps involve creating a cell graph that is subsequently refined and rendered more accurate via a network denoising process. Building on this refined graph, scGAC employs a graph attentional autoencoder to capture the nuanced representations of cells which are ideal for clustering. This phase facilitates information flow across cells, each given variable weightings, effectively discerning latent intercellular relationships. To bring the process full circle, scGAC utilizes a self-optimizing strategy in order to acquire cell clusters.
Diving deeper into the model, ScGAC's first step encompasses the creation of a cell graph wherein nodes represent individual cells and the edges, carry weightings determined through the Pearson correlation coefficient. To enhance the integrity of this initial graph, any superfluous or inconsistent edges are meticulously pruned through network denoising. With this streamlined graph in place, ScGAC harnesses a graph attentional autoencoder to learn latent cellular representations, seamlessly integrating insights from gene expression profiles and the intricate intercellular connections. The attention mechanism, enables ScGAC to distribute distinct weights to  cellular neighbors as information cascades through the cellular network. In the culmination of this process, ScGAC seamlessly blends representation learning with clustering optimization. Through the self-regulating mechanism, the system evaluates similarities between cells and their respective clustering centroids using a defined membership matrix, Qc×n. Memberships are subsequently adjusted via an optimized matrix, Pc×n. Qc×n is then employed to recalibrate clustering centroids and deliver the final, refined clustering outcome.

Another GNN-based model proposed for single-cell data clustering is SpaDAC (SPAtially embedded Deep Attentional graph Clustering) \cite{huo2023integrating}. It has been introduced as a revolutionary technique to harness multi-modal data, pinpointing spatial domains and concurrently reconstructing denoised gene expression profiles. At its core, SpaDAC is adept at learning low-dimensional embeddings from spatial transcriptomics data. It achieves this by creating multi-view graph modules, meticulously designed to encapsulate both spatial location interconnections and morphological ties. Rather than relying on singular data sources, SpaDAC synthesizes joint representations, melding complementary multimodal insights. Central to its functionality, the method incorporates an attention mechanism, expertly capturing the similarities among neighboring spatial spots. Moreover, to further refine and bolster the compact nature of these latent representations, SpaDAC incorporates an iterative self-optimizing clustering module, ensuring heightened precision in the clustering outcomes.

In the same vein, scASGC \cite{wang2023scasgc} has been proposed as an unsupervised clustering technique derived from an adaptive, streamlined graph convolution model. This innovative approach meticulously constructs credible cell graphs, aggregates information from neighboring nodes using the streamlined graph convolution model, and adaptively ascertains the optimal count of convolution layers tailored to diverse graphs. Initially, the similarity matrix for cells is computed, and, leveraging Network Enhancement (NE), extraneous noise within the similarity matrix is filtered out to establish a trustworthy undirected cell graph. To bolster both the efficiency and precision of the model, a streamlined graph convolution model is employed, integrating the principles of the multi-head attention mechanism. This fosters a heightened discernment of intricate cell-cell interactions and fosters a richer comprehension of latent cell attributes. Consequentially, a dynamic approach is harnessed to determine the ideal quantity of convolution layers for varying graphs, followed by the application of spectral clustering. 

In addition, GLAE \cite{shan2023glae} is an advanced graph-based auto-encoder tailored for scRNA-seq analysis, distinguished by its ability to extract cell relation graphs from a multitude of perspectives and leverage them for cell clustering; in contrast to traditional GNN-based approaches that rely on static graph inputs, GLAE emphasizes dynamic learning, allowing cell relation graphs to evolve and be refined with every iteration, offering a more robust framework compared as noise is filtered during the graph learning process; a hallmark of GLAE is its unique modular structure, built to understand cell relation graphs from varied angles in parallel; to address specific challenges, cell relation graphs are reconstituted at each stage based on the gene expression data, facilitating an adaptive update of the graphs throughout training; furthermore, the system employs several parallel sub-modules to examine discrete gene subsets and produce cell relation graphs from diverse viewpoints while the intricate relationship network between gene subsets is discerned through cell embeddings originating from these sub-modules; essentially, GLAE's design focuses on the generation and application of cell relation graphs for optimal cell clustering and subsequent analyses.

GLAE facilitates the extraction of M cell relationship graphs through M individual sub-modules. Within the confines of the gene subset relationship network learning module, a specific network corresponding to gene subsets is derived, subsequently guiding the aggregation of cell relationship graphs and cell embeddings. Each sub-module, encapsulates three pivotal elements: (1) gene subsets learning, (2) cell relationship graph construction, and (3) feature encoding and decoding phases. Each sub-module crafts a relationship graph of cells, which integrates chosen features and acquires data from its immediate predecessor. Cell samples, denoted as X, undergo processing in M sub-modules, leading to the construction of M cell relationship graphs each providing a unique viewpoint. These M graphs, along with the cell embeddings, are chaneled to the gene subset relationship network learning module. Herein, a particular gene subset network emerges and subsequently participates in the aggregation module, creating cell relationship graphs and cell embeddings. In essence, GLAE is partitioned into four distinct sectors: (1) cell relationship graph extraction from multiple aspects, (2) crafting the gene subset relationship network, (3) aggregating cell relationship graphs and cell embeddings, and (4) clustering of cells.

scTAG \cite{yu2022zinb} is a another pioneering method that blends the functionalities of a deep graph convolutional network to learn cell–cell topological representations and pinpoint cell clusters in a cohesive process; it seamlessly combines the zero-inflated negative binomial (ZINB) model within a topology adaptive graph convolutional autoencoder, achieving a compact latent representation; furthermore, the method leverages the Kullback–Leibler (KL) divergence to optimize clustering. An innovative aspect of scTAG lies in its simultaneous optimization for clustering loss, ZINB loss, and cell graph reconstruction loss, enabling the combined optimization of cluster labeling and feature learning, all while preserving the intricate topological structures. In essence, through the construction of a cell graph, scTAG is adept at maintaining both the intrinsic topological information and intricate cell-cell relations inherent in scRNA-seq datasets.

Moreover scGPCL \cite{lee2023deep}, is introduced as a prototypical contrastive learning method underpinned by graph techniques for scRNA-seq data clustering. This approach primarily harnesses Graph Neural Networks to encode cell representations in a cell–gene graph, preserving the relational information intrinsic to scRNA-seq data. To further refine cell representations, scGPCL employs prototypical contrastive learning, a strategy designed to separate dissimilar pairs semantically while drawing closer the similar ones. 

This technique diverges from traditional practices in the domain, where cells are interconnected in a cell–cell graph based on pre-established cell similarities; instead, scGPCL creates a bipartite cell–gene graph, marking connections between cells and genes when specific genes are expressed in designated cells, as evidenced by the provided input gene expression matrix.

ScGMM-VGAE \cite{lin2023scgmm} makes use of a Gaussian mixture model-based variational graph autoencoder specifically for scRNA-seq data, achieving enhanced cell clustering. By integrating a statistical clustering model with a deep learning algorithm, the system employs a graph variational autoencoder (VGAE) to process both a cell-cell graph adjacency matrix and a gene feature matrix. This produces latent data which is subsequently clustered by the Gaussian mixture model (GMM) module. A unique loss function, incorporating parameter estimates from the GMM and VGAE, is instrumental in the optimization stage.
In the field of high-dimensional scRNA-seq data, the scGMM-VGAE framework specializes in cell clustering. The GMM module takes on the task of clustering cell types from VGAE-encoded latent data. This interaction ensures that GMM parameters significantly influence the VGAE's learning process. Taking the VGAE's low-dimensional representation into account, the probabilistic GMM model navigates the distribution. As a result, latent features which vividly capture the intrinsic characteristics of scRNA-seq data become the cornerstone of the cell clustering process.

\subsection{Dimensionality reduction}


Given the intrinsic nature of single-cell sequencing as a high-throughput technology, it inherently generates datasets characterized by a vast number of dimensions pertaining to both cells and genes. This multifaceted complexity leads to a phenomenon known as the 'curse of dimensionality,' which is pervasive in single-cell RNA sequencing (scRNA-seq) data. The manifestation of this challenge is further complicated by the fact that not all genes are equally informative or relevant in the context of cell clustering based on expression profiles. While efforts have already been directed towards the mitigation of this issue through feature selection, it represents merely a preliminary step. The next phase involves the utilization of specialized dimensionality reduction algorithms. Serving as a crucial component in the preprocessing stage, these algorithms function to reduce data complexity, thereby facilitating subsequent visualization and analysis \cite{heumos2023best}.

Traditional, non-machine-learning-based tools, such as t-SNE \cite{van2008visualizing} and UMAP \cite{mcinnes2018umap}, are commonly employed in the field, possibly owing to their relatively straightforward implementation and interpretability. However, these conventional algorithms often encounter limitations when tasked with deciphering interpretable dimensionality reduction from a high-dimensional gene space that is dense and complex. In an effort to circumvent these constraints, deep-learning-based dimensionality reduction techniques have emerged in recent years \cite{lopez2018deep}, \cite{risso2018general}. Despite this advancement, these deep learning methodologies have typically focused solely on embedding unique cellular features, neglecting to consider the intricate relationships between individual cells. Such an oversight restricts their capacity to unravel the complex topological structure inherent among cellular entities, consequently hindering the analysis of developmental trajectories. GNN-based models offer a promising alternative, demonstrating substantial potential by preserving long-distance relationships within the latent space. This sophisticated approach leverages interconnected network structures, thereby offering a more nuanced understanding of data, and possibly providing insights into previously obscure cellular relationships \cite{luo2021topology}.

In this vein, CellVGAE \cite{buterez2022cellvgae} is a model developed to harness the capabilities of graph neural networks for the unsupervised exploration of scRNA-seq data. Constructed upon the foundation of a variational graph autoencoder (VGAE) architecture enhanced with graph attention layers (GAT), this model is distinctively designed to function on the direct connectivity among cells, placing a significant emphasis on dimensionality reduction and clustering.

Unlike traditional neural networks that primarily source information from gene expression values, CellVGAE capitalizes on cell connectivity, depicted as a graph, as its inductive bias. This facilitates the execution of convolutions on non-Euclidean structures, aligning seamlessly with the geometric deep learning paradigm. For operational efficiency, CellVGAE incorporates k-nearest neighbor (KNN) alongside Pearson correlation graphs, commonly referred to as PKNN, recognized for their proficient performance and extensive application within the field.

Similarly, scGAE (Single-Cell Graph Autoencoder) \cite{luo2021topology} has been introduced as a dimensionality reduction technique adept at retaining the topological structure inherent in scRNA-seq data. scGAE creates a cell graph and employs a multitask-centric graph autoencoder, ensuring simultaneous preservation of both topological structure details and feature information of scRNA-seq data. This methodology has been further adapted to cater to visualization, clustering, and trajectory inference in scRNA-seq data.

The model combines the strengths of both deep autoencoders and graph-based models, facilitating the embedding of the intricate topological framework of high-dimensional scRNA-seq data into a more low-dimensional latent space. Subsequent to procuring the normalized count matrix, scGAE formulates an adjacency matrix amid cells utilizing the K-nearest-neighbor algorithm. Through graph attentional layers, the encoder transitions this count matrix into a low-dimensional latent space. scGAE then interprets the embedded data via both a feature decoder and a graph decoder. While the feature decoder's objective is the reconstruction of the count matrix to maintain feature information, the graph decoder aims to recover the adjacency matrix, ensuring retention of topological structure details. The embedded data is subsequently decoded to the same dimensions with the original dataset, targeting a minimized distance between the input and the reconstructed data. For the dual objectives of learning data embedding and cluster designation, deep clustering is employed.

SCDRHA \cite{zhao2021scdrha} is structured around two key components: the Deep Count Autoencoder (DCA) dedicated to data denoising and a graph autoencoder, designated for data projection into a low-dimensional latent space. The graph autoencoder, constructed on the Graph Attention Network (GAT) framework, aims to project the data into a latent expression while preserving cellular topological structures. Given that the graph autoencoder receives single-cell graphs as input, consisting of node matrices and an adjacency matrix, the adjacency matrix—crafted via the K-nearest-neighbor (KNN) algorithm—holds significant implications. 

However, due to the pronounced sparsity of scRNA-seq data, the KNN algorithm can distort the adjacency matrix. To address the challenges presented by dropout events on the KNN algorithm's output, their impact has been underscored and the robust denoising capabilities of DCA to counteract zero inflation resulting from such dropouts have been leveraged. As both the original and the DCA-reconstructed data share the same dimensionality, an initial dimensionality reduction for the latter is executed using Principal Component Analysis (PCA). Employing the latent space delineated by PCA, the graph autoencoder operates to further reduce dimensions, creating a low-dimensional embedding conducive for both visualization and clustering. The primary model, DCA, is tailored to counteract dropout events and is trained via a ZINB model-based autoencoder. The subsequent model, a GAT-based graph autoencoder, serves to map the DCA-denoised data into a low-dimensional latent space. The process initiates by normalizing raw data, followed by data denoising using DCA. The culmination of the process witnesses the integration of the PCA-compressed matrix and the adjacency matrix, fed into the GAT-based graph autoencoder, yielding a low-dimensional embedding.

\subsection{Cell annotation}
Cell annotation, refers to advanced computational methodologies, that facilitate the efficient labeling of individual cells or cell clusters through the integration of algorithmic processes and existing biological knowledge. By identifying specific gene expression signals or signatures within a cell or cluster that correspond to recognized patterns of known cell types or states, the targeted cell or cluster can be accordingly labeled. This technique is a part of the broader framework of cell type annotation, a process that assigns identities to cells or clusters based on their gene expression profiles. The intricate resolution of cellular heterogeneity across diverse tissues, developmental stages, and organisms achieved through this approach enhances our understanding of cellular dynamics and gene functionality. This comprehensive perspective not only elucidates fundamental biological processes but also offers critical insights into the underlying mechanisms governing both healthy physiological states and pathological conditions \cite{clarke2021tutorial}. Despite continuous advancements, the effectiveness of existing cell identification/annotation methods remains constrained, a limitation that can be attributed in part to the incomplete utilization of higher-order cellular characteristics. In response to these inherent challenges, research is progressively turning to deep learning models, including Graph Neural Networks (GNNs). These advanced computational frameworks are uniquely capable of evaluating both low-order and high-order data features, synthesizing them through specialized network topologies.

In this vein, SCEA \cite{abadi2023optimized} represents a novel clustering methodology tailored for scRNA-seq data. It leverages two distinct components for dimensionality reduction coupled with a self-optimizing mechanism for cell annotation. This involves an initial application of a multi-layer perceptron (MLP)-based encoder followed by a GAT – graph attention auto-encoder-. By integrating two sequential units, SCEA is able to obtain both accurate cell and gene embeddings.

The SCEA algorithm encompasses several stages: a) preprocessing of input data, b) graph creation and denoising, c) dimensionality reduction, and d) data clustering using the K-means algorithm. A graph is initially constructed utilizing the Pearson's correlation coefficient method. Subsequent to this, graph pruning is facilitated using the Network Enhancement (NE) method. This method deploys a doubly stochastic matrix to identify and eliminate noisy edges. Notably, a matrix is deemed doubly stochastic only when all its entries are non-negative, and the summation of elements in each row and column equals one. Among non-negative matrices, stochastic and doubly stochastic matrices are distinguished by several unique properties. Dimension reduction is then achieved in two stages: the initial phase employs an encoder based on the MLP architecture, and subsequently, a graph attention autoencoder utilizes a cell graph to further reduce the dimensionality of the encoder's output. By capitalizing on the denoised cell graph that captures cell connectivity information, the graph attention autoencoder discerns cellular relationships, enhancing the overall clustering outcome.

In addition, HNNVAT \cite{wang2023adversarial} presents an advanced adversarial dense graph convolutional network architecture tailored for single-cell data classification. Central to its design, the architecture takes advantage of the integration of a dense connectivity mechanism and attention-based feature aggregation to augment the representation of sophisticated higher-order features, while fostering a seamless combination amongst them for feature learning in convolutional neural networks. A distinctive feature reconstruction module, is incorporated to ensure the fidelity of the original data's features, fortifying its primary classification objective. Additionally, HNNVAT harnesses the potential of virtual adversarial training, amplifying its generalization capabilities and robustness. The processed single-cell data is channeled into a composite structure merging both a fully connected network and a graph convolutional neural network. This incorporation of dense connectivity and attention-based aggregation is pivotal for enhancing higher-order feature representation. An encoder-decoder based feature reconstruction module is seamlessly integrated to uphold the intrinsic features of the original dataset.

HNNVAT commences with data preprocessing, composed of two parts: filtering cells and atypical genes, and feature selection. Leveraging the processed data, an adjacency matrix is crafted, containing the top 1,000 genes for each respective dataset. Non-diagonal elements embody interactions among distinct genes, sourced from the STRING database. This process culminates in a weighted adjacency matrix, underpinned by the probabilities of gene interactions. When geared towards single-cell classification, the convolutional neural network’s input is a matrix with gene expression values. Architecturally, HNNVAT is composed of four key modules: a hybrid neural network, an attention-based convolutional hierarchical aggregation, local feature reconstruction, and virtual adversarial training. The model's initiation involves inputting both the gene expression matrix and the gene adjacency matrix into the four-layer graph convolutional neural network, so that node feature extraction can be performed. Recognizing the inherent risk of model overfitting from the extracted high-order features, direct connections are strategically positioned between hidden layer neurons and each subsequent convolution layer. This ensures that insights gleaned from nodes in ensuing layers are harnessed to further refine the model.

Another GNN-based cell annotation tool is scDeepSort \cite{shao2020reference}. It is a cutting-edge cell-type annotation tool tailored for single-cell transcriptomics, powered by a deep learning model based on a weighted graph neural network (GNN); its efficacy has been highlighted through rigorous testing on both human and mouse scRNA-seq datasets, confidently annotating a massive collection of 764,741 cells spanning 56 human and 32 mouse tissues; driven by the innate graph structure inherent to scRNA-seq data, where cells express genes, scDeepSort leverages this dynamic, basing its training on comprehensive single-cell transcriptomic atlases: the human cell landscape (HCL) and mouse cell atlas (MCA); each cell's unique graph network, which encompasses its genes and neighboring cells, plays a pivotal role in the supervised learning of scDeepSort using known cell labels extracted from these atlases for every tissue. 

scDeepSort, is comprised of three pivotal components: an embedding layer that retains the graph node representations (remaining frozen during training), a weighted graph aggregator layer adept at learning graph structural details and creating a linear feature space for cells (with a modified GraphSAGE framework acting as the backbone GNN), and, last but not least, a linear classifier layer tasked with categorizing the final cell state representation into one of the established cell type classes. 

scAGN \cite{bhadani2023attention} employs an attention-based graph neural network, specifically designed for detecting cell types within scRNA-seq datasets through label-propagation. This approach leverages transductive learning, a unique technique where both training and testing datasets are involved during the learning process. By examining patterns across the combined datasets, the model identifies patterns that are then utilized to predict labels for the unlabeled test data points.

The unique attention-based graph neural network, stands out by eliminating intermediate fully-connected layers. Instead, propagation layers are replaced by attention mechanisms, ensuring the integrity of the graph structure is maintained. This sophisticated attention system enables an adaptive, dynamic summarization of local neighborhoods, paving the way for more precise predictions. When applying the scAGN technique, single-cell omics data are inputted post batch-correction, utilizing both canonical correlation analysis and mutual nearest neighborhood (CCA-MNN). Through the power of transductive learning, scAGN can derive cell labels for query datasets using reference datasets with pre-established or expert-annotated labels.

In addition, scMRA \cite{yuan2022scmra} introduces a knowledge graph that embodies the unique traits of cell types present in multiple datasets. This graph then pairs with a graphic convolutional network, which acts as a distinguishing mechanism. The aim of scMRA is twofold: to preserve the inherent closeness within cell types and to maintain the spatial positioning of these cell types throughout various datasets. By combining multiple datasets, a meta-dataset emerges, forming the basis for annotation, and as an added bonus, scMRA has the prowess to account for and eliminate batch effects.

scMRA is capable of harnessing insights from several well-documented datasets and applying this information to unlabelled target data. The training part for this model is broken down into four pivotal stages. Initially, sequencing data, plagued by dropout events during the sequencing process, are aligned. This alignment is achieved by leveraging a ZINB model-based denoising autoencoder. Following this, the next step involves computing a prototype representation for each cell type across all datasets. This global prototype is then updated through a moving average strategy, which aids in tempering the inherent randomness of these calculations. It also highlights the variability of identical cell types when viewed across different datasets. Building on this, the third stage sees the creation of a knowledge graph based on the global prototype. This graph becomes a comprehensive source of information, with the weight of connections between two prototypes being contingent on their mutual resemblance. The knowledge graph  is further broadened by integrating samples from a query batch. Finally, a graphic convolutional network (GCN) is deployed. Its primary role is to efficiently relay feature representations across the expanded graph, leading to the derivation of a classification probability for every individual node.

ScGraph \cite{yin2022scgraph} is another cutting edge cell annotation algorithm that harnesses gene interaction relationships to augment cell-type identification efficacy. It employs a graph neural network to aggregate information from interacting genes. By, utilizing the gene interaction network it mitigates technical disruptions and autonomously discern cell types. Integration of gene expression and gene interaction data enables ScGraph not only to determine the cell type of specific cells but also to highlight vital gene interaction relationships from experimental data. Training on the Human Cell Landscape (HCL) dataset allowed ScGraph to identify cell types in a distinct human scRNA-seq dataset using the acquired model, showcasing its aptitude for precise cell-type identification using a reference dataset. ScGraph capitalizes on gene interaction relationships to accumulate neighboring information for each gene, subsequently enhancing cell embedding and identification. To gauge ScGraph's performance with diverse foundational networks, seven human gene interaction networks and one mouse gene interaction network were examined.

In essence, ScGraph is a graph neural network that receives scRNA-seq data and a gene interaction network as its inputs to automatically infer cell labels. It is structured into three distinct modules: (i) a graph representation module, (ii) a feature extraction module, and (iii) a classification module. The gene interaction network can be naturally visualized in a graph where each node represents a gene and edges denote the relationships between them. Within the graph representation module, constructed as a single graph convolutional layer, every node's data is refined by aggregating information from its adjacent nodes. This module incorporates a modified GraphSAGE convolutional layer for enhanced representation.

\subsection{GRN inference}

Gene expression, a cornerstone of cellular functions, is regulated through intricate gene relations, the understanding of which can reveal potential drug targets. These interactions can be visualized through a gene regulatory network (GRN), where nodes represent genes and directed edges signify causal gene-to-gene effects. Traditionally, GRN inference relied on analyzing steady-state data from gene knockout experiments, observing alterations in gene expression upon silencing specific genes \cite{aalto2020gene}. However, the emergence of single-cell multi-omics technologies has ushered in advanced computational techniques that utilize genomic, transcriptomic, and chromatin accessibility data, offering substantial precision in GRN inference \cite{badia2023gene}. Existing tools have faced challenges in inferring the dynamic biological networks present across varied cell types and their reactions to external stimuli. Viewing supervised GRN inference through the lens of a graph-based link prediction problem offers a promising approach, emphasizing the generation of vector gene representations to predict potential regulatory interactions. Graph neural networks, tailored for network representation generation, demonstrate exceptional efficacy in GRN inference even with limited cell samples.

For instance, GENELink \cite{chen2022graph} has been introduced to deduce latent interactions between transcription factors (TFs) and their respective target genes within gene regulatory networks (GRN). This is achieved by utilizing a graph attention network (GAT) framework. GENELink adeptly maps single-cell gene expressions associated with observed TF-gene pairs into a latent space. Subsequently, distinct gene representations are learned for the purpose of downstream similarity evaluations or causal inference between gene pairs by refining the embedding space. Utilizing a GAT-based supervised framework, the GRN is inferred by deriving low-dimensional vector representations from single-cell gene expressions, given an incomplete prior network. Intriguingly, GAT integrates self-attention mechanisms within graph neural networks, adapting them to directed graphs, making them viable for both transductive and inductive applications. The supervised GRN inference challenge is regarded as a link prediction task: with a predetermined set of genes and some of their known interactions, the goal is to ascertain other concealed connections within the network. GENELink harnesses GAT to cohesively assimilate gene interactions from prior knowledge with gene expression data sourced from scRNA-seq experiments.

The model is architecturally designed with a dual GAT layer structure followed by MLPS to learn low dimentional gene representations. The multi-layer perceptrons (MLPs) further refine low-dimensional gene representations, facilitating subsequent similarity measurements or causal inferences between genes. Multi-head attention mechanisms are employed to model regulatory intensity and refine node representations. GENELINK requires concurrent input of single-cell gene expression data and a prior adjacency matrix. Throughout the message-passing phase, attention mechanisms are leveraged to assign implicit weights to neighboring nodes. Additionally, two GAT layers are incorporated to identify higher-order neighbor data. Post GAT layers, pairwise genes, designated as I and j, are independently inputted in two channels. Both comprise multi-layered neural networks, generating dense gene representations. The dot product is employed as the scoring mechanism to measure the similarity between genes i and j.

Similarly, DeepMAPS \cite{ma2023single} presents a sophisticated approach to biological network inference from scMulti-omics, interpreting this data through a heterogeneous graph. By leveraging both local and global contexts, it discerns intricate relationships among cells and genes with the aid of a multi-head graph transformer. The underlying strength of DeepMAPS lies in its foundation: the heterogeneous graph transformer (HGT). This model creates a heterogeneous graph where cells and genes are represented as nodes and their relationships serve as edges. Notably, it captures both immediate and global topological attributes to deduce cell-cell and gene-gene relationships alike. Additionally, its built-in attention mechanism evaluates the relevance of genes to specific cells, enhancing the depth of gene contribution information. Unlike many traditional models, HGT operates without binding to pre-existing assumptions, allowing it the flexibility to highlight often overlooked gene regulatory relationships.
The methodology of DeepMAPS is both intricate and comprehensive. Initial steps involve preprocessing of the data, with emphasis on discarding low-quality cells and genes with low expression values. Once filtered, data is normalized, leading to the creation of an integrated cell-gene matrix that captures the collective activity of genes within individual cells. From this matrix, a heterogeneous graph is created where cells and genes are represented as nodes and edges represent the existence of genes in individual cells. This sets the stage for the HGT model to determine low-dimensional embeddings for cells and genes, while also identifying the importance of genes in relation to cells through the use of attention scores. Based on these insights, DeepMAPS then proceeds with predicting cell clusters and functional gene groupings. Ultimately, the system infers a diverse array of biological networks specific to each cell type.

Moreover, DeepTFni \cite{li2022inferring} has been designed to infer Transcriptional Regulatory Networks (TRNs) from single-cell assays that assess transposase-accessible chromatin through sequencing (scATAC-seq) datasets. With the integration of a graph neural network, optimized for network representation, DeepTFni displays superior efficacy in TRN deduction, even when confronted with a sparse cell count. It should also be noted that, DeepTFni's application unveiled important Transcription Factors (TFs) that play a central role in tissue evolution and tumorigenesis. It has been highlighted that numerous genes associated with mixed-phenotype acute leukemia underwent a significant transformation within the TRN, even when the changes in messenger RNA levels remained relatively stable.

DeepTFni harnesses the capabilities of VGAE to calculate latent embeddings and the holistic topology of a graph. Throughout it’s operation, the TRN (Transcriptional Regulatory Networks) is visualized as an undirected graph $G \epsilon \{V, E\}$, with nodes (V) representing TFs and edges (E) indicating their interactions. DeepTFni's input is a scATAC-seq count matrix, which is then processed to yield the imputed TRN. Viewing TRN inference as a link-prediction problem, DeepTFni operates in three stages. Initially, a TRN skeleton is formulated, acting as an incomplete prior. This structure encompasses a set of TF pairs that have a maximum likelihood of exhibiting regulatory interactions. This preliminary blueprint not only offers a genuine benchmark for discerning new TF interactions during model training but also stands as the ground truth during inference. This initial phase is propelled by a meticulous examination of accessible TF gene promoters, and interactions are determined based on TF motif occurrence in another TF's accessible promoter. Subsequently, the TRN blueprint is represented through the use of an adjacency matrix. In the subsequent phase, the node feature, represented by a regulatory potential (RP) score, is computed using scATAC-seq data. This RP score, drawn from the MAESTRO methodology, mirrors the aggregate regulation imposed by adjacent scATAC-seq peaks on a specific gene within a cell. In the final stage, the VGAE model is constructed with an encoder, constituted by a two-layer graph convolution network, and a decoder, which employs an inner product succeeded by a logistic sigmoid function. The encoder processes the initial adjacency matrix and the node feature matrix, yielding latent representations for each TF node. These latent representations are then transformed into TF interactions via the decoder. To minimize any inadvertent effects of the stochastic model elements, a ten-pass prediction strategy has been implemented, ensuring the robustness of TF interactions. The final output is the imputed TRN, which is represented by the restructured adjacency matrix.

\subsection{Cell-cell interactions}
Cell-cell interactions (CCIs) are pivotal for the seamless functioning and development of multicellular organisms. These interactions, ranging from physical connections to intricate communication pathways, allow cells to transmit vital information to one another \cite{liu2022evaluation}. For instance, one cell might release specific molecules, such as growth factors, which then bind to receptors on neighboring cells, resulting in changes in gene expression. With the advent of single-cell RNA sequencing (scRNA-seq) technology, scientists now have a powerful tool to delve deeper into these interactions at a granular level, offering computational insights into how individual cells communicate and function within a complex system \cite{xie2023comparison}. Most cell-cell interaction inference methodologies, however, frequently fall short in capturing the convoluted interplay within cells, and the overarching influence of pathways and protein complexes. To navigate this complexity and harness the rich relationships between cells, ligands, receptors, and various annotations, the advent of graph-based deep learning has proven to be effective for cell-cell interaction inference. 

For example, GraphComm \cite{so2023graphcomm} is introduced as a novel graph-based deep learning approach tailored for the prediction of cell-cell communication (CCC) within single-cell RNAseq datasets. Through the leveraging of an established model and subsequent refinement of a network based on single-cell transcriptomic data, GraphComm proficiently forecasts CCC activity spanning various cells. Furthermore, it discerns its impact on downstream pathways, spatially proximate cells, and alterations resultant from drug interventions. Central to its functionality, GraphComm harnesses ligand-receptor annotations, encompassing protein complex data and pathway details, alongside expression values. This combination facilitates the crafting of cell interaction networks. Employing feed-forward methodologies, it endeavors to master an optimal data representation, subsequently predicting the likelihood of CCC interactions. GraphComm can allocate numerical values that elucidate the interrelationships amongst cell clusters, ligands, and receptors. This capability empowers the extraction of communication probabilities associated with specific ligand-receptor connections, rendering a hierarchical ranking of CCC activities.

In its methodology, GraphComm makes use of a scRNAseq dataset paired with a meticulously curated ligand-receptor database, initiating the formulation of a directed graph that mirrors the inherent CCC ground truth. Augmenting this with both protein complex data and pathway annotations, a Feature Representation technique is employed to extract the spatial information of ligands and receptors within this directed graph. Subsequently, GraphComm creates a secondary directed graph that delineates the associations between cell groups and ligands/receptors. This graph is enriched with transcriptomic details, cell group insights, and positional attributes sourced from the prior Feature Representation step, resulting in the acquirement of updated numerical node attributes through a Graph Attention Network. These Node Features can be strategically employed through inner product operations to calculate communication probabilities for all feasible ligand-receptor pairings. When these calculated probabilities, derived from the Graph Attention Network, are integrated with the secondary directed graph, it facilitates the establishment of ligand-receptor connections characterized by top-tier CCC activity. This sophisticated model provides a foundation for visually representing activities at both the ligand-receptor and cell group levels.

Similarly, spaCI \cite{tang2023spaci}, is an adaptive graph-based deep learning model equipped with attention mechanisms, seeks to interpret cell-to-cell interactions derived from SCST profiles. This model integrates both the spatial positioning and gene expression profiles of cells, aiming to highlight the active L–R signaling axis among adjacent cells. Crucially, spaCI is adept at recognizing the upstream transcriptional factors that facilitate these active L–R interactions.

spaCI's methodology capitalizes on the inherent spatial connections between cells, combined with the expression data from all genes to identify the L–R interactions. Within the framework of spaCI, genes are projected into a latent space through a dual-component system: a gene-centric linear encoder and a cell-centric attentive graph encoder. This dual approach ensures the inclusion of both gene expression patterns and spatial cellular associations into the latent space. To guarantee distinct delineation between interactive and non-interactive pairs, spaCI employs a triplet loss mechanism.

Another GNN-based model for cell-cell interaction inference is scHGA (single-cell heterogeneous graph cross-omics attention model)  \cite{liu2022heterogeneous}. scHGA is grounded in a heterogeneous graph neural network that incorporates two attention mechanisms. It facilitates a unified analysis of single-cell multi-omics data sourced from diverse protocols, such as SNARE-seq, scMT-seq, and sci-CAR. To circumvent the inherent heterogeneity in single-omics data, scHGA adeptly learns a cell association graph to extract neighboring cell information. The hierarchical attention mechanism allows for the calculation of a latent representation of aggregated cells, thus integrating knowledge across diverse omics. This, in turn, aids in deconstructing cellular heterogeneity and offers an enhanced methodology to categorize cellular characteristics. Evidently, scHGA signifies a profound application of graph neural networks for the analysis of single-cell multi-omics, shedding fresh light on the comprehension of single-cell sequencing datasets. 

The model receives a preprocessed single-cell transcriptome expression matrix and a single-cell epigenomic peak matrix as foundational input. It combines bipartite graphs of multi-omics to create a heterogeneous graph. Subsequent to this, cell neighbors within single-omics data are aggregated via cell similarity derived from a meta-path, referred to as node-level attention. Acknowledging the varied significance of different omics in cellular representation, the model aggregates multi-omics cell neighbors via cross-omics attention, resulting in the final latent representation. Such a structured approach empowers scHGA to adeptly highlight cellular heterogeneity, reinforcing relations between cells.

\subsection{Other applications}

It is imperative to underscore that beyond the aforementioned six primary categories, GNN-based strategies have been also been proposed for specialized analyses in other realms of single-cell omics data. For instance, OmicsGAT \cite{baul2022omicsgat} is a novel graph attention network (GAT) proposed for cancer patient stratification as well as disease phenotype prediction. The model designed to adeptly blend graph-based learning with a dedicated attention mechanism for RNA-seq data analysis. Within its architecture, the multi-head attention mechanism distinguishes itself by proficiently extracting information from specific samples. This is achieved by distributing varied attention coefficients to the adjacent samples. Detailed experiments on The Cancer Genome Atlas (TCGA) datasets, including breast and bladder cancer bulk RNA-seq data, coupled with two single-cell RNA-seq datasets, yielded two essential conclusions. First, OmicsGAT effectively incorporates neighborhood data of a particular sample and meticulously calculate an embedding vector. This tailored vector considerably enhances disease phenotype predictions, refines stratifications among cancer patients, and optimizes cell sample clustering. Second, the attention matrix, derived from the multi-head attention coefficients, presents a rich depth of valuable information. Impressively, this matrix surpasses the capabilities of standard sample correlation-based adjacency matrices.

When examining a particular sample, each of the h heads in the network allocates distinct attention coefficients to the adjacent samples. Among these, the most prominent coefficient is selected for each neighboring sample, indicating its association with the target sample. Adopting this method, a full attention matrix is constructed. This result highlights the intrinsic value of the attention mechanism, especially when coupled with traditional insights provided by the adjacency matrix. Within the OmicsGAT framework, an embedding is derived from gene expression datasets. This methodology is grounded in the notion that samples, whether they represent patients or cells, exhibiting similar gene expressions are likely to demonstrate analogous disease trajectories or cell types, forming inherent interconnectedness. Some neighboring samples might exert a more significant influence on subsequent predictions or cluster formations, a subtlety that standard similarity metrics may overlook. Addressing this, OmicsGAT dynamically adjusts attention levels across neighbors of a sample for a singular head during the embedding generation. To capture diverse insight from neighbors and ensure a robust learning process, this strategy is mirrored across multiple heads, encapsulating many independent attention mechanisms within a multi-head model.

Moreover, STAGATE \cite{dong2022deciphering} is a sophisticated graph attention auto-encoder designed to adeptly pinpoint spatial domains. Its proficiency stems from its ability to learn low-dimensional latent embeddings by integrating spatial information with gene expression profiles. Notably, to delineate the spatial similarity at the boundaries of these domains, STAGATE incorporates an attention mechanism. This mechanism is calibrated to adaptively discern the similarity between adjacent spots, and the system can be further enhanced with a cell type-aware module, which is achieved by integrating gene expression profile pre-clustering. A significant feature of STAGATE is its adaptability to span multiple consecutive sections. This capability ensures a reduction in batch effects between sections and enables the extraction of three-dimensional (3D) expression domains from the reconstructed 3D tissue. Substantial tests across diverse ST data generation platforms, including 10x Visium, Slide-seq, and Stereo-seq, underscored STAGATE's superiority, especially for tasks like spatial domain identification, data visualization, spatial trajectory inference, data denoising, and 3D expression domain extraction.

Diving into its operational mechanics, STAGATE begins by creating a spatial neighbor network (SNN) grounded on the relative spatial positioning of spots. If needed, a cell type-aware SNN can be integrated by refining the SNN using pre-established gene expression clusters. This pre-clustering approach adeptly pinpoints areas with distinct cell types, enabling the modified SNN to finely map the spatial similarity at the boundaries of these unique spatial domains, a feature particularly vital for ST data with low spatial resolutions. Sequentially, STAGATE employs a graph attention auto-encoder to derive low-dimensional latent embeddings, combining both spatial metrics and gene expressions. Each spot's normalized expression undergoes a transformation into a d-dimensional latent embedding via an encoder, only to be reverted into a reconstructed expression profile through a decoder. Distinct from traditional auto-encoders, STAGATE introduces an attention mechanism within the middle layer of the encoder-decoder architecture. This innovation empowers STAGATE to adaptively learn the edge weights of SNNs, refining the similarity metrics between neighboring spots. These weights are subsequently harnessed to refresh the spot representation by collectively aggregating data from its immediate neighborhood. The resultant latent embeddings set the stage for data visualization using tools like UMAP, and the identification of spatial domains is achieved through clustering algorithms, exemplified by mclust and Louvain.

In the field of DEG (Differentially Expressed Genes) identification Cellograph has been proposed \cite{shahir2023cellograph}. It is a semi-supervised framework that leverages graph neural networks to assess the effects of perturbations at the single-cell level. Notably, Cellograph quantifies the typicality of individual cells in various conditions and constructs a latent space tailored for interpretive data visualization and clustering. A byproduct of its training process, the gene weight matrix, reveals crucial genes that differentiate between conditions. Grounded in Graph Convolutional Networks (GCN) – a graph counterpart to the conventional CNNs – Cellograph emphasizes node classification. It navigates through scRNA-seq data sourced from diverse conditions, perceiving individual cells as nodes. Through its two-layer GCN, Cellograph captures the latent embedding of single-cell data, identifying the representativeness of each cell relative to its baseline sample label. 

This latent space fosters adept clustering, grouping cells with similar treatment responses and transcriptomic signatures. Additionally, this space can be conveniently projected into two dimensions for insightful visualizations. Benchmarking against existing methods has proven that, Cellograph excels in delineating perturbation effects and presents a distinctive GNN framework for single-cell data visualization and clustering.
The methodological approach involves transforming single-cell data, gathered from varied drug treatments, into a kNN graph. In this structure, cells act as nodes, while edges connect transcriptionally similar cells. This kNN graph is then processed through a two-layer GCN, enabling Cellograph to both quantitatively and visually determine how each cell aligns with its experimental label, using latent embeddings and softmax probabilities. Once created, the latent space becomes compatible with k-means clustering techniques. As a final step, a gene-centric weight matrix is employed to spotlight genes that are instrumental in distinguishing between different conditions.

STitch3D \cite{wang2023construction} is computational framework designed to combine multiple 2D tissue slices, enabling the reconstruction of 3D cellular structures that span individual tissues to entire organisms. By merging models of these 2D tissue slices with cell-specific expression profiles procured from single-cell RNA sequencing, STitch3D effectively highlights 3D spatial regions characterized by consistent gene expression, while also unveiling the spatial distribution of various cell types. Notably, STitch3D can distinguish inherent biological variability across slices from potential batch effects, capitalizing on the common information between slices to construct robust 3D tissue models. It receives inputs as spatially-determined gene expression matrices from various ST slices, paired with specific cellular gene expression profiles from an associated scRNA-seq dataset. By analyzing these inputs together, STitch3D identifies 3D spatial domains marked by consistent gene expressions and discerns the spatial distribution of refined cell types. Key operations of STitch3D include configuring 3D spatial coordinates by aligning several tissue slices and generating a global 3D neighborhood graph. Building upon these inputs, STitch3D employs a deep learning model that ensures combination of information across tissue slices, subsequently aiding in the detection of 3D spatial regions and decomposition of cell types. One of STitch3D's primary mechanisms is its encoder network, tailored to transform spatial spots from diverse slices into a common latent space. Following this, it deploys a neural network termed the deconvolution network to deduce cell-type proportions from these latent spatial representations. This specific network is optimized to replicate initial raw ST gene expressions by merging the anticipated cell-type ratios with the cellular gene expression profiles from the scRNA-seq reference.

Inputs to the model comprise of raw datasets from numerous ST tissue sections and cell-specific gene expression profiles from a reference scRNA-seq dataset. During the pre-processing phase, spots from various tissue slices are aligned to create 3D spatial coordinates, succeeded by the creation of a global 3D graph. The primary architecture in STitch3D integrates these structures to facilitate representation learning, focusing on the identification of 3D spatial territories and cell-type deconvolution. Outputs from STitch3D include the identified 3D spatial domains and spatial cell type distributions estimation in tissues. Additionally, it offers pathways for advanced analyses such as spatial trajectory prediction, refining of low-quality gene expression measurements, crafting virtual tissue slices, and recognizing genes marked by 3D spatial expression patterns. In its approach, STitch3D processes slices collectively and employs a graph attention-based neural network to extract latent representations of spots, integrating them with 3D spatial data. Augmenting its capabilities, STitch3D integrates a graph attention network, specifically designed to encompass the established 3D spatial graph. This network consists of a graph attention layer and a dense layer, with the attention layer processing inputs like a global adjacency matrix and normalized, log-transformed gene expressions. While the input dimensionality corresponds to a select number of highly variable genes—a metric that is dataset-dependent—the output dimensionality remains a constant 512.

PIKE-R2P \cite{dai2021pike} is a distinctive method designed for single-cell RNA to protein prediction. This approach integrates both protein–protein interactions (PPI) and prior knowledge embedding within a graph neural network framework. When provided with a sample of scRNA-seq data, PIKE-R2P predicts the abundance levels of various proteins. The model is chiefly divided into two components: a PPI-driven GNN and an embedding of prior knowledge.

The first component incorporates the PPI-based graph neural network into the dataset. These interactions offer a channel for information transfer among proteins, signifying the collective promotion of specific biological tasks, such as mutual inhibition or promotion. This is is achieved by encoding the PPIs within a graph structure, with nodes symbolizing proteins and edges depicting interactions. Consequently, the graph neural network calculates the outcome of this information exchange stemming from interactions amongst proteins. The secondary component focuses on embedding established knowledge, which encompasses factors like co-expression and gene co-occurrence. Given the conservation of PPI relationships across various cell types, databases such as STRING, with extensive PPI data, are harnessed for this knowledge embedding process.

In addition, scFEA \cite{alghamdi2021graph}, is a computational framework, that deduces single-cell fluxome from single-cell RNA-sequencing (scRNA-seq) data. Drawing strength from a meticulously restructured human metabolic map, segmented into focused metabolic modules, it integrates a unique probabilistic model. This model is fine-tuned to apply flux balance constraints on scRNA-seq data, paired with an advanced graph neural network-based optimization solver. The method seamlessly captures the transition from transcriptome to metabolome via multi-layer neural networks, ensuring the non-linear dependency between enzymatic gene expressions and reaction rates is maintained. Key innovations within scFEA address prevailing challenges: a probabilistic model harnessed to apply the flux balance constraint across diverse metabolic fluxomes for numerous single cells; a strategic approach to reducing metabolic map size, considering both network topology and gene expression status; a multi-layer neural network model designed to trace the dependency of metabolic flux on enzymatic gene expressions; and a graph neural network framework and solution, maximizing the overall flux balance of intermediary substrates across all cells.

Three pivotal computational components define scFEA: network reorganization, individual cell metabolic flux estimation, and a series of downstream analyses. These analyses encompass metabolic stress estimation, metabolic gene perturbation, and cell clustering based on varied metabolic states. The method transforms the entire metabolic network into a factor graph, comprising sets of metabolic modules and flux balance constraints. Network reduction occurs by examining network topology, metabolic gene expression status, and optional customized network regions. In the subsequent phase, each module's metabolic flux becomes a model, portraying a non-linear function of the enzyme expression levels within the module. This non-linearity is discerned by a 2-4 layer neural network. For neural network parameter resolution, scFEA introduces a flux balance constraint interlinking the modules of all individual cells in the tissue, all based on a probabilistic model. Finally, scFEA carries out comprehensive downstream analysis.

Moreover, GLUE (graph-linked unified embedding) \cite{cao2022multi} represents a modular framework designed for the seamless integration of unpaired single-cell multi-omics data while also inferring regulatory interactions. Through modeling of these regulatory interactions across omics layers, it adeptly accounts for the discrepancies among diverse omics-specific feature spaces, doing so in a manner that aligns with biological intuition. When tested on intricate tasks such as triple-omics integration, integrative regulatory inference, and the construction of a multi-omics human cell atlas spanning millions of cells, GLUE exhibited proficiency in rectifying prior annotations. Cell states within this framework are conceptualized as latent cell embeddings obtained via variational autoencoders. Due to the inherent disparities in their biological characteristics and assay methodologies, individual omics layers are each paired with a distinct autoencoder. This specific autoencoder employs a probabilistic generative model that is meticulously adapted to the feature space unique to that layer.

Within its architecture, GLUE employs omics-specific variational autoencoders to distill low-dimensional cell embeddings $U1, U2, U3$ from every omics layer. While the data dimensionality and generative distribution may vary between layers, a consistent embedding dimension, denoted as m, is maintained. For the purpose of cohesively connecting the omics-specific data spaces, GLUE harnesses prior insights regarding regulatory interactions, encapsulated in a guidance graph $G=(V,E)$. Here, the vertices $V=V1 \cup V2 \cup V3$ signify omics features. A graph variational autoencoder is then tasked with deriving feature embeddings $V=(V\tau1,V\tau2,V\tau3)$ from the guidance graph. These feature embeddings are subsequently utilized by data decoders, which, through an inner product with cell embeddings, reconstruct omics data. This process adeptly bridges omics-specific data spaces, ensuring a harmonized embedding orientation. Concluding this intricate process, an omics discriminator D is employed to synchronize the cell embeddings across the various omics layers, utilizing adversarial learning techniques.

Furthermore, The single-cell Graph Convolutional Network (scGCN) \cite{song2021scgcn} founded on a graph-based artificial intelligence framework has been introduced to enable dependable and replicable integration as well as proficient knowledge transfer across varied, distinct datasets. When evaluated against multiple label transfer methodologies using an expansive collection of 30 single-cell omics datasets, spanning diverse tissues, species, sequencing platforms, and molecular strata such as RNA-seq and ATAC-seq, scGCN has consistently exhibited enhanced accuracy.

In its operational framework, scGCN initially learns a hybrid, sparse graph containing both inter-dataset and intra-dataset cell mappings. This is achieved by utilizing mutual nearest neighbors of canonical correlation vectors, which coherently project the diverse datasets into a common low-dimensional space. This methodical approach fosters identification and dissemination of mutual information between reference and query datasets. Following the graph's creation, a semi-supervised GCN is employed to align cells from both the reference and subsequent datasets onto a common latent space. This ensures that cells sharing labels are located within a homogenous cluster. As a result, labels within the query dataset are effectively ascertained and assimilated from the reference dataset.

GrID-Net \cite{wu2023econometric} is a neural network based on the time-tested Granger causal inference, promoting graph-based evaluations over conventional sequential analyses. This algorithm has seen applications in the domain of human corticogenesis single-cell studies, providing predictions for neuronal cis-regulatory elements and facilitating the interpretation of genetic variants related to schizophrenia (SCZ). It draws upon the econometric principle of predictive temporal causality, where tissue-specific locus–gene associations are identified, whereby the accessibility of a given locus acts as a precursor to the gene's expression. The underlying time disparity between chromatin accessibility and gene expression culminates in the “cell-state parallax”. This concept captures the temporal offset between epigenetic and transcriptional events stemming from their intrinsic cause-and-effect dynamics. During evolving biological processes, the chromatin regulatory potential stands as a tangible representation of this parallax, suggesting the potential to discern such phenomena from single-cell snapshots.

Extracting insights from this parallax through single-cell instances aids in the derivation of causal connections between loci and genes. GrID-Net, when applied to bimodal single-cell datasets—co-assaying both chromatin accessibility (ATAC-seq) and gene expression (RNA-seq)—seeks to delineate the associations between noncoding regions (or peaks) and the genes they regulate. Notably, GrID-Net can extend beyond traditional Granger causality to incorporate DAG-based renderings of single-cell trajectories, anchored by its core mechanism: a graph neural network that adopts a lagged message propagation structure. GrID-Net, thus, is adept at identifying asynchronous dynamics between regulatory activities at accessible noncoding sites and their subsequent influence on gene expression. Furthermore, the peak-gene associations discerned through GrID-Net assist in the functional dissection of genetic variations linked with pathological conditions.

For ligand-receptor identification, GCNG (Graph Convolutional Neural Networks for Genes) \cite{yuan2020gcng} has been proposed. It is a model that adeptly translates spatial attributes into a graph representation, subsequently integrating it with expression datasets through supervised learning. Not only does GCNG enhance spatial transcriptomics data analysis methodologies, but it also identifies novel extracellular gene interaction pairs. Furthermore, the outcomes generated by GCNG are primed for subsequent analytical tasks, such as functional gene assignment. In order to predict extracellular interactions derived from gene expression using GCN, spatial transcriptomics data is initially transmuted into a graph that represents the interactions between cells. Following this, each gene pair's expression is encoded and processed by GCNG, blending the graph data with the expression datasets through convolution. This methodology empowers the neural network to harness not only first-order relationships but also more intricate higher-order interactions inherent in the graph's architecture.

GCNG focuses on deducing gene interactions pivotal for intercellular communications from spatial single-cell expression datasets. The model receives two inputs: the cellular locations in spatial images and the gene pair expressions within these cells. The initial process within GCNG encompasses the conversion of single-cell spatial expression data into two distinct matrices. The first matrix captures cellular positions in the form of a neighborhood graph, while the second matrix encodes gene expressions within individual cells. These matrices collectively serve as the foundational input for a five-layered graph convolutional neural network, centered on the prediction of intercellular gene communication relationships. 

An instrumental component of the GCN architecture is the graph convolutional layer, facilitating the combination of the graph's structure (spanning cell locations and neighboring relations) with node details (pertaining to gene expression in a designated cell). Given that the graph structure associates spatially adjacent cells, GCNs can draw upon the convolutional layers without requiring any direct image data. To delve deeper into the GCNG structure, it encompasses two graph convolutional layers, a singular flatten layer, a dense layer with 512 dimensions, and a classification-centric sigmoid function output layer. The inclusion of a dual convolutional layer set-up equips the system to comprehend indirect graph relationships. Recognizing that regulatory proteins can wield influence beyond their immediate neighborhood, this methodology enables the inference of interactions potentially overlooked when merely accounting for adjacent entities.

Another cutting edge GNN-based tool for single-cell analysis is StdGCN \cite{li2023spatial}. StdGCN is an advanced graph neural network model meticulously designed for the deconvolution of cell types within spatial transcriptomic (ST) data. This model harnesses the computational prowess of graph convolutional networks (GCN), a deep learning model that predominantly employs graph-based architecture. It adeptly utilizes the rich reservoir of single-cell RNA sequencing (scRNA-seq) data as a reference point. Notably, STdGCN pioneers the integration of expression profiles derived from single-cell datasets with the spatial positioning details procured from ST data, enhancing the precision of cell type deconvolution.

The operation of STdGCN begins with the identification of marker genes distinctive to particular cell types, coupled with the generation of pseudo-spots, drawing upon the scRNA-seq datasets. Sequentially, the model creates two linking graphs to represent the GCN workflow. The first link graph termed the expression graph, is a composite entity, composed of three sub-graphs: an internal pseudo-spots graph, an internal real-spots graph, and a graph bridging real to pseudo-spots. The genesis of each sub-graph is firmly rooted in mutual nearest neighbors (MNN), harnessing the expression-based similarities between spots. The second link graph, termed the spatial graph, is created from the Euclidean distances discerned between real-spots within ST datasets. During StdGCN’s operation, the input feature matrix is propagated through both the expression GCN layers as well as it’s spatial layers. The resultant outputs,  Exp-feature and Spa-feature, are subsequently concatenated column-wise, into a unified matrix. This matrix then undergoes processing through fully-connected layer, yielding predictions pertaining to cellular type proportions for each specific spot. For model training, pseudo-spots undergo a division into training and validation subsets. It is imperative to note that only the training subset of pseudo-spots is utilized for back propagation, while the validation subset is strategically reserved for early-stopping protocols. Adopting this methodology, real-spot cellular type proportions are updated through the GCN framework, further enhancing the learning process of the pseudo-spots.

Last but not least, SCAN-IT \cite{cang2021scan} has been proposed to effectively translate the spatial domain identification problem into an image segmentation problem. In this process, cells emulate pixels, while the expression values of genes encapsulated within a cell take on the role of color channels. The foundation of SCAN-IT is anchored in geometric modeling, graph neural networks, and an informatics strategy known as DeepGraphInfomax. Notably, SCAN-IT exhibits remarkable versatility, accommodating datasets spanning an array of spatial transcriptomics methodologies. This includes datasets characterized by heightened spatial resolution but low gene coverage and those marked by reduced spatial resolution but augmented gene coverage.

Delving into the intricacies of tissue segmentation, deep graph neural networks are harnessed, culminating in a reimagining of the process as a clustering-oriented image segmentation task. Within this paradigm, spatial transcriptomics (ST) data is envisioned as an image distinguished by irregular grid structures and thousands of channels. SCAN-IT, a sophisticated deep learning algorithm, is introduced to impartially discern tissue domains from such images. In its initial phase, SCAN-IT meticulously constructs a geometry-aware spatial proximity graph, with the nodes representing either the spots (cells groups) or the individual cells present in the ST dataset, employing the alpha complex methodology. When compared to the conventionally employed k-nearest neighbor graphs, this unique graph representation offers a more nuanced reflection of the physical closeness between cells. 

\begin{figure}[!h]
    \centering
    \begin{adjustbox}{width=\textwidth,center}
        \includegraphics[width=\textwidth]{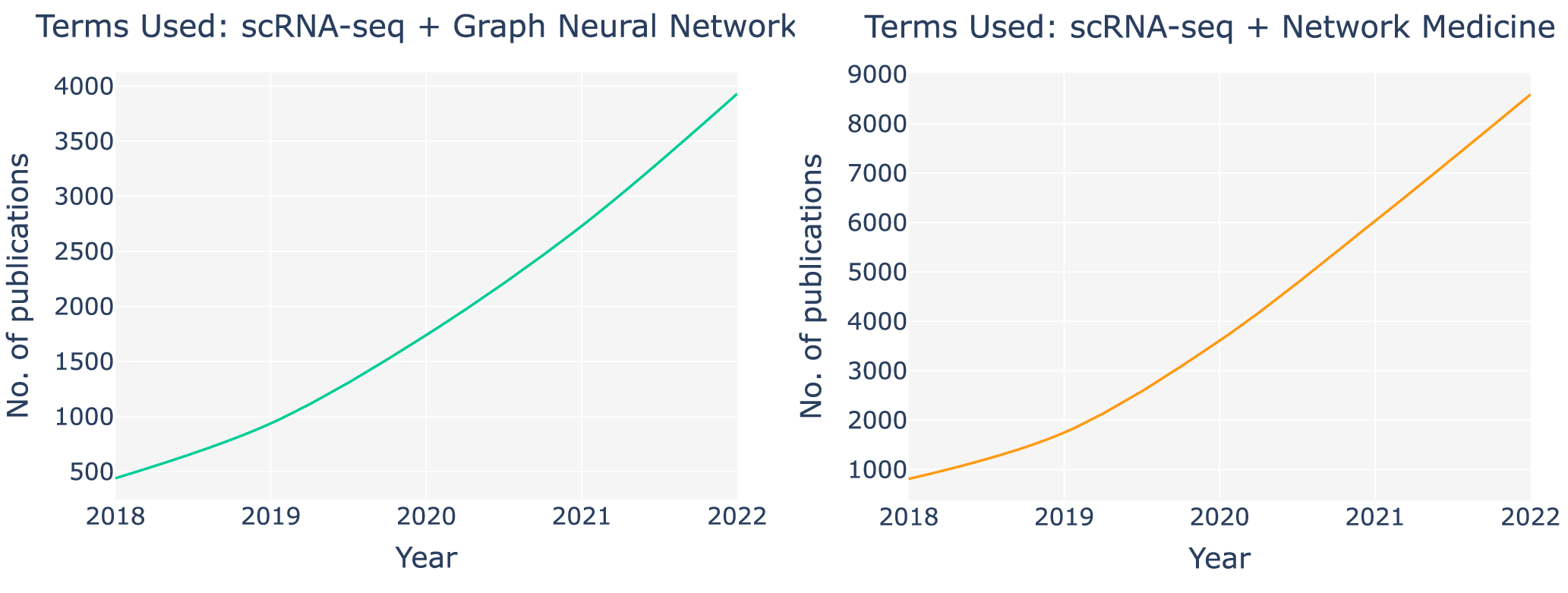}
    \end{adjustbox}
    \caption{Line-plots depicting the number of papers published from 2018 to 2022. The terms used for the first search were scRNA-seq and Graph Neural Network. The terms used for the second search were scRNA-seq and Network Medicine. Through these plots the increasing popularity of network medicine and graph neural network frameworks for single-cell data analysis is highlighted. The term search took place through the use of google scholar.}
    \label{fig:your_label}
\end{figure}

Adapting the DeepGraphInfomax, low-dimensional node embeddings are produced. Intriguingly, these embeddings encapsulate gene expression trajectories inherent to the cells, as well as the gene expression pathways manifest in their immediate microenvironment. This encapsulation is facilitated by the inherent attributes of DeepGraphInfomax which prioritize neighborhood information conservation. The culmination of this process witnesses the transformed low-dimensional data being channelled into established clustering algorithms, facilitating the delineation of tissue domains.

\section{Discussion \& Future Perspectives }

Remarkably, in the last four years, 39 GNN tools tailored for single-cell data have been recorded, with almost all published in high-impact journals.  It should be noted that 16 out of the 39 tools make use of attention mechanisms, a trend that underlines the increasing potential and adaptability of such models in capturing intricate relationships within single-cell data \cite{chen2022graph, buterez2022cellvgae, feng2022single, xu2021efficient, baul2022omicsgat, cheng2022scgac, huo2023integrating, dong2022deciphering, feng2023single, abadi2023optimized, wang2023scasgc, zhao2021scdrha, so2023graphcomm, wang2023construction, tang2023spaci, bhadani2023attention}. Moreover, the majority of these tools, with few exceptions -only 6 to be precise- rely on single-cell RNA-sequencing expression data as their primary input, highlighting the diverse array of single-cell data forms gaining traction \cite{huo2023integrating, dong2022deciphering, tang2023spaci, li2022inferring, yuan2020gcng, cang2021scan}.

The surging interest in GNNs for single-cell omics data analysis is unsurprising given their inherent dual strengths. Firstly, their foundation on deep neural networks enables such models to inherently tackle the challenges posed by big data \cite{zhang2018survey}. As single-cell datasets continue to grow both in size and complexity, such capabilities become indispensable. Secondly, GNNs by nature operate on graphs, an unrivaled advantage in capturing complex relationships and interconnectivities between entities. With the expanding heterogeneity of omics data, such as the growing popularity of single-cell ATAC and spatial data which provide crucial insights to understanding the complexities of intercellular communication present in both healthy and pathological tissues through precise localization of gene clusters expressed in specific cell subpopulations as identified through scRNA-seq \cite{longo2021integrating} the importance of GNNs becomes even more pronounced. Thus not only are they enable to integrate multi-omics data but also to preserve the intricate relationships between cells and genes, a task made feasible primarily through the use of graphs.

Diving deeper into the evolution of the GNNs, it's evident that innovation remains relentless. Since their inception, GNNs have been continually refined and augmented. Graph Convolutional Neural Networks (GCNs) \cite{kipf2016semi} represented a monumental stride, which was later complemented and extended by the introduction of Graph Attention Networks \cite{velivckovic2017graph}. A fairly recent breakthrough in this lineage is the introduction of GATV2 \cite{brody2021attentive}. GATv2 emerges as a refined iteration of Graph Attention Networks, addressing the constraints of static attention observed in its predecessors. This notion of static attention is typified by a homogenized attention ranking for pivotal nodes across all querying nodes. GATv2 introduces an innovative dynamic attention dimension, which endows the model with the agility to modulate attention weightings contingent on the query node in question. This dynamism is achieved via a recalibrated attention coefficient derivation mechanism, wherein node embeddings are synergistically concatenated, subsequently undergoing a non-linear Leaky ReLU activation function. The emergent output engages in a dot product interaction with a mutable weight vector. This strategic shift enables GATv2 to adeptly map the graph's anatomy and accentuate pivotal node dialogues by endowing it with a dynamic attention weighting mechanism.

Summarizing, we estimate that the revolutionary potential of GNNs in the field of single-cell omics is undeniable.  Their inherent capabilities, combined with the ever-evolving architectures, paint a promising picture for the future. Beyond the evident applications, there lies the fascinating prospect of GNNs paving the way for groundbreaking advancements in drug repurposing as well as precision and network medicine. In fact, as highlighted in \textbf{Fig.2}, there has been a notable surge in publications over the past 5 years that combine scRNA-seq with either GNNs or the broader field of network medicine. By doing so, GNNs have the potential to spearhead the development of innovative diagnostic tools and therapeutic strategies for a plethora of complex diseases, further solidifying their central role in the field of biomedicine.

\section*{Declarations}
\subsection*{Conflict of Interest}
The authors have no competing interests to declare that are relevant to the content of this article.
\subsection*{Data availability}
No datasets were generated or analysed during the current study.

\bibliography{sn-bibliography}

\end{document}